\documentclass[journal, table, xcdraw, twoside]{IEEEtran}

\IEEEoverridecommandlockouts
\usepackage{adjustbox}
\usepackage{amsmath,amssymb,amsfonts}
\usepackage{algorithmic}
\usepackage{color}
\usepackage{environ}
\usepackage{float}
\usepackage{graphicx}
\usepackage[utf8]{inputenc}
\usepackage{lipsum}  
\usepackage{listings}
\usepackage{multirow}
\usepackage[numbers, square, comma,compress]{natbib}
\usepackage{pgfplots}
\usepackage{relsize}
\usepackage{siunitx}
\usepackage{subcaption}
\usepackage{textcomp}
\usepackage{tikz}
\usepackage{tikz-3dplot}
\usepackage{balance}
\usepackage{url}
\usepackage{fancyhdr}
\usepackage[normalem]{ulem}
\usepackage[pdftex]{hyperref}
\usepackage{caption}
\DeclareMathOperator{\st}{subject \ to:} 

\DeclareUnicodeCharacter{2212}{−}
\usepgfplotslibrary{groupplots,dateplot}
\usetikzlibrary{patterns,shapes.arrows, matrix}
\pgfplotsset{compat=newest}

\NewDocumentCommand{\evalat}{sO{\big}mm}{%
  \IfBooleanTF{#1}
   {\mleft. #3 \mright|_{#4}}
   {#3#2|_{#4}}%
}

\newcommand{\vect}[3]{{_{\mathsmaller{\mathrm{#2}}}\mathbf{#1}_{\mathsmaller{\mathrm{#3}}}}} %
\newcommand{\wfr}[0]{\ensuremath{W}} %
\newcommand{\bfr}[0]{\ensuremath{B}} %

\newcommand{\bm}[1]{\boldsymbol{#1}}

\newcommand{\rom}[1]{(\expandafter{\romannumeral #1\relax})}

\newcommand{\comment}[1]{}

\begin{document}

\title{
Performance, Precision, and Payloads: \\
Adaptive Nonlinear MPC for Quadrotors
}

\author{Drew Hanover, 
        Philipp Foehn,
        Sihao Sun, 
        Elia Kaufmann,
        Davide Scaramuzza
        \thanks{Manuscript received: September, 8, 2021; Accepted November, 10, 2021.}
        \thanks{
        This paper was recommended for publication by Editor Pauline Pounds upon evaluation of the Associate Editor and Reviewers' comments.
        The authors are with the Robotics and Perception Group, Dep. of Informatics, University of Zurich, and Dep. of Neuroinformatics, University of Zurich and ETH Zurich, Switzerland (\protect\url{http://rpg.ifi.uzh.ch}).
        This work was supported by the National Centre of Competence in Research (NCCR) Robotics through the Swiss National Science Foundation (SNSF) and the European Union’s Horizon 2020 Research and Innovation Programme under grant agreement No.~871479 (AERIAL-CORE) and the European Research Council (ERC) under grant agreement No.~864042 (AGILEFLIGHT).
        }
        \thanks{Digital Object Identifier (DOI): 10.1109/LRA.2021.3131690}
}

\maketitle

\begin{abstract}
Agile quadrotor flight in challenging environments has the potential to revolutionize shipping, transportation, and search and rescue applications.
Nonlinear model predictive control (NMPC) has recently shown promising results for agile quadrotor control, but relies on highly accurate models for maximum performance. 
Hence, model uncertainties in the form of unmodeled complex aerodynamic effects, varying payloads and parameter mismatch will degrade overall system performance.
In this paper, we propose $\mathcal{L}_1$-NMPC, a novel hybrid adaptive NMPC to learn model uncertainties online and immediately compensate for them, drastically improving performance over the non-adaptive baseline with minimal computational overhead. 
Our proposed architecture generalizes to many different environments from which we evaluate wind, unknown payloads, and highly agile flight conditions.
The proposed method demonstrates immense flexibility and robustness, with more than 90\% tracking error reduction over non-adaptive NMPC under large unknown disturbances and without any gain tuning.
In addition, the same controller with identical gains can accurately fly highly agile racing trajectories exhibiting top speeds of 70 km/h, offering tracking performance improvements of around 50\% relative to the non-adaptive NMPC baseline.
\end{abstract}

\vspace{-8pt}
\section*{Supplementary Material}

\noindent Video: \protect\url{https://youtu.be/8oB1rG5iYc4} \\

\vspace{-8pt}

\section{Introduction}\label{sec:introduction}

\subsection{Motivation}
Unmanned aerial vehicle (UAV) utilization in industrial applications is increasing at an astounding rate~\cite{Loianno2020jfr,air_taxi_2020, Ackerman2018Skydio}.
There are over 300,000 commercial drones registered in the US alone according to the Federal Aviation Administration~\cite{FAA2021UAS}.
Significant commercial opportunities exist for robust autonomous systems which can safely conduct inspection of sensitive, hazardous, and remote systems under uncertainty. 
As a result, 
the global drone service market is expected to grow from 4.4 Billion USD to 63.6 Billion between 2018 and 2025~\cite{newswire_2019}.
In transportation and shipping applications, UAVs can improve efficiency of operations via autonomous missions and high speed maneuvers leading to dramatic time savings, cost reductions, and higher throughput~\cite{foehn2021cpc}.
On a similar note, emergency scenarios raised the need for agile autonomous systems to conduct search and rescue tasks when time is of the essence~\cite{Faessler16jfr}.
\begin{figure}[!t]
    \centering
       \begin{subfigure}[b]{0.5\textwidth}
   \includegraphics[width=0.98\textwidth,trim={200 400 200 400},clip]{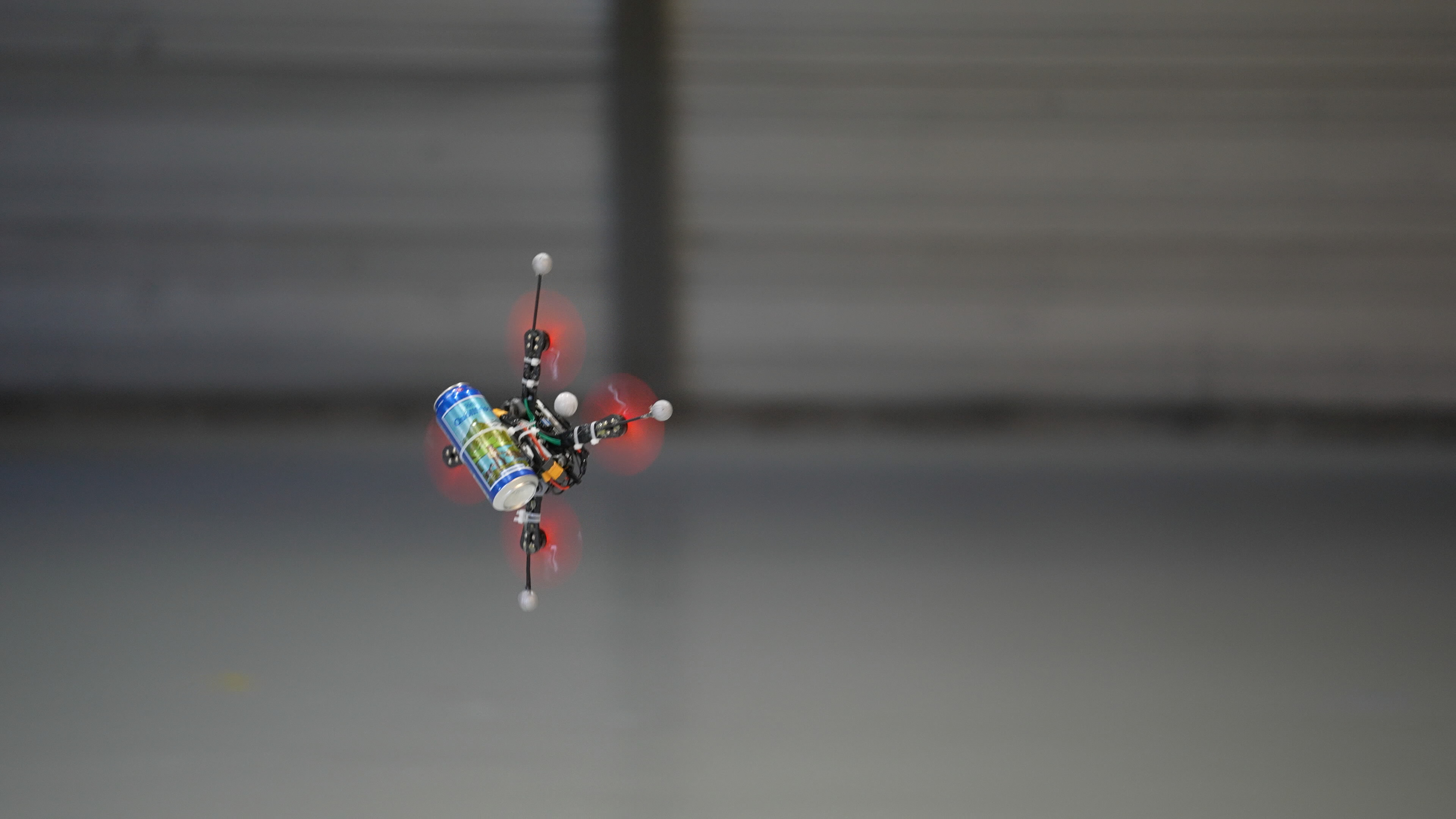}
\end{subfigure}
   \begin{subfigure}[b]{0.5\textwidth}
    \includegraphics[width=0.98\textwidth]{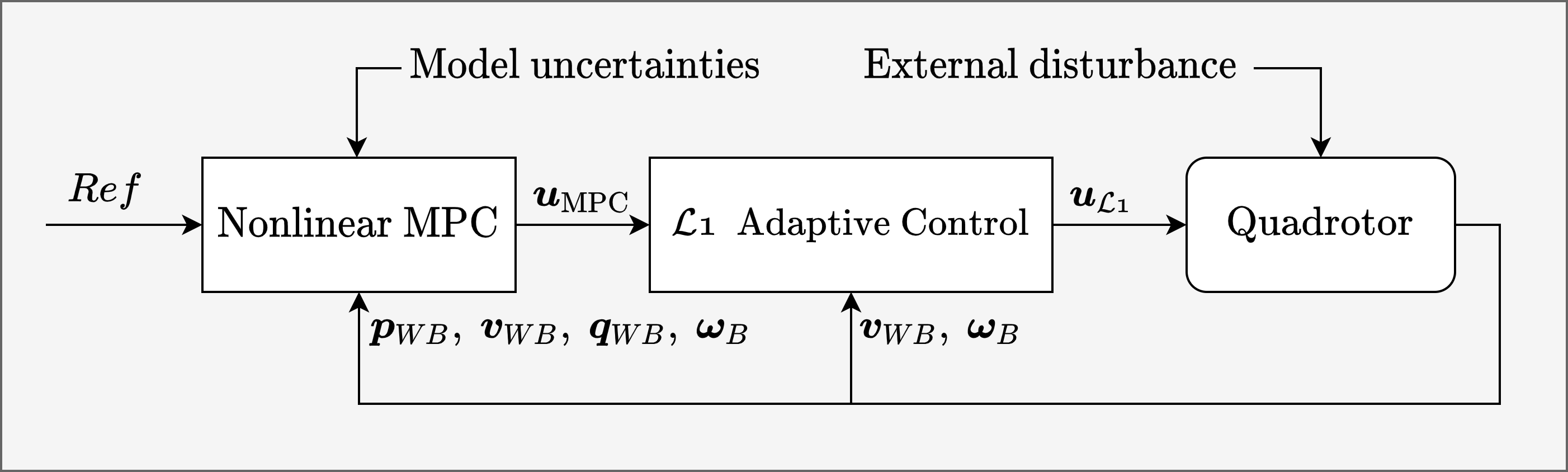}
    \end{subfigure}
    \caption{Top: Quadrotor carrying a beer payload of unknown mass while flying an aggressive racing trajectory.
    Our proposed $\mathcal{L}_1$-NMPC method allows to reduce tracking errors on such agile trajectories by more than 90\% compared to non-adaptive methods.
    Bottom: $\mathcal{L}_1$-NMPC control diagram.}
    \label{fig:Nearmiss}    
\end{figure}
To facilitate these opportunities, UAVs must be able to accurately track agile trajectories in presence of model uncertainties and external disturbances such as unknown drag coefficients, varying payloads, or wind gusts respectively.
These uncertainties can significantly degrade the performance and reliability of the system and potentially lead to loss of control if not compensated for. 
High-fidelity physics-based models can improve control performance, but are often prohibitively expensive to procure and require extensive levels of domain expertise~\cite{mehndiratta2020gaussian, torrente2021datadriven, sun2019quadrotor, bauersfeld2021neurobem}.
With the advancements of data-driven methods such as those described in~\cite{mehndiratta2020gaussian, torrente2021datadriven, sun2019quadrotor}, the costs of obtaining accurate models has been dramatically reduced.
Models learned from data, however, have a tendency to overfit and can be intractable to update online~\cite{bauersfeld2021neurobem}.

In reality, modeling the system with complete accuracy is impossible, necessitating control algorithms which are robust to uncertainty.
The field of robust control has tried to address this problem with varying degrees of success. 
Classical methods such as $H_{\infty}$ and more recently stochastic model predictive control may enable safe behavior of the autonomous system, but at the cost of significant performance degradation~\cite{zhou1998essentials,petersen2012robust}.
Predicting external disturbances such as wind-gusts is often impractical due to the chaotic nature of the disturbance~\cite{sun2019quadrotor, Nisar2019ral}.
This necessitates the development of highly adaptive control algorithms which can robustly compensate for unknown model dynamics without prior assumptions, while pushing the system to its dynamic limits.

\subsection{Contribution}
In this work, we propose a novel quadrotor control architecture which cascades a nonlinear model predictive controller (NMPC) to an $\mathcal{L}_1$ adaptive controller to fly highly aggressive racing trajectories under various types of model uncertainties and external disturbances with speeds up to \SI{20}{\meter\per\second}.
We show that the addition of the adaptive controller can drive the real system towards the behavior specified by the underlying MPC model with significantly better tracking performance when compared to a non-adaptive baseline. 
The proposed controller can compensate in real-time, while being transferable to many different applications such as flying in windy environments, carrying completely unknown payloads, and flying aggressive racing trajectories without any re-tuning of control gains.
When unknown payloads up to 60\% of the quadrotor mass are introduced into the system, our approach demonstrates a reduction of tracking error over 90\% compared to non-adaptive NMPC methods onboard a real quadrotor.
Additionally, we show that it is even possible to accurately fly racing trajectories with unknown slung payloads.
The data indicates we can track these trajectories \emph{with} an unknown payload representing 13\% of the quadrotor's mass with 44\% higher accuracy than a non-adaptive NMPC \emph{without} a payload attached.
Our experiments under nominal model conditions demonstrate at least a 50\% tracking performance improvement over state of the art data-driven MPC methods \emph{without} an aerodynamics model on a set of increasing speed circle trajectories which exhibit speeds of up to 36~km/h, indicating that the adaptive component is able to learn and compensate for unmodeled aerodynamic effects in real time with minimal computational overhead.

\section {Related Work}
Trajectory tracking controllers for quadrotors have been studied extensively over the last decade.
A detailed survey on quadrotor control methods can be found in~\cite{Nascimento2019MAVControlSurvey, sun2021comparative, romero2021model}. 
Most of the existing literature focuses on hover conditions or slow speed maneuvers which satisfy small angle assumptions necessary for linear control methods.
We are interested in exploring a wider flight envelope, and focus on advanced techniques to push the physical limits of the platform. 

Agile flight of aerial vehicles has been a top priority for the aerospace industry for the better part of 80 years, as a part of which
NASA developed Model Reference Adaptive Control (MRAC) to deal with large model uncertainties that are difficult to model and measure \cite{mareels1987mitRule}. 
Readers interested in the underlying mathematics of adaptive control methods for aerospace vehicles are pointed to the following references~\cite{lavretsky2013robust, aastrom2013adaptive, ioannou2012robust, hovakimyan2010L1}.
Specifically, we focus on the $\mathcal{L}_1$ adaptive control approach due its inherent ability to provide rapid adaptation that is decoupled from the robustness of the controller~\cite{cao2008designL1}. 
Applications of $\mathcal{L}_1$ adaptive controllers have been successfully demonstrated across a variety of aerial vehicles such as fixed-wings, quadcopters, and octocopters~\cite{beard2006l1, gregory2009l1,mallikarjunan2012l1}.  
The main feature of the $\mathcal{L}_1$ adaptive controller is to drive a system towards a desired reference model behavior. 
Typically, this is done using a linear reference model to specify the desired behavior, however this can lead to unrealistic desired dynamics which cannot be achieved by the real system. 
Adaptive control has successfully demonstrated accurate trajectory tracking using quadrotors in several works~\cite{dydek2013adaptiveuav, nicol2011robustadaptivecontrol, Schreier2012modelingandadaptivecontrol, kotaru2020geometric}.
However, the maneuvers conducted are typically simple step inputs, or slow speed circles which do not exploit the inherent agility of the quadrotor platform. 

Adaptive controllers often act as an augmentation to an existing baseline controller rather than as a standalone controller. 
Authors in~\cite{bujarbaruah2018adaptive} take advantage of the high level planning of linear MPC, cascaded with an adaptive control law to adapt to persistent model mismatch. 
A cascaded linear MPC with a linear reference model $\mathcal{L}_1$ adaptive controller was demonstrated for the quadrotor trajectory tracking problem with exogenous disturbances in~\cite{pereida2018adaptivempc}, but the trajectories demonstrated were simplistic and slow. 
They claim a reduced computational cost compared to nonlinear MPC frameworks, however the limitations of linear optimal control when applied to nonlinear problems are well understood, especially when the vehicle exhibits highly aggressive maneuvers~\cite{KAMEL2017linVsNonLinMPC}. 

Similarly, in~\cite{pravitra2020l1adaptive} a Model Predictive Path Integral (MPPI) controller was coupled with a nonlinear reference model $\mathcal{L}_1$ adaptive controller for agile quadrotor flight, however the authors have not shown feasibility of the proposed method on real hardware.
No analysis of the adaptive control signal is provided, however video footage released of the simulation performance of the proposed $\mathcal{L}_1$-MPPI architecture indicates highly oscillatory control performance in the first-person camera view\footnote{https://youtu.be/f602VSGIVb0}.
Additionally, the high level MPPI controller can only run at a rate of 50 Hz using desktop hardware which makes it infeasible for on-board control. 

Previous works demonstrating agile trajectory tracking with physical quadrotors include~\cite{tal2020accurate,torrente2021datadriven,faessler2017differential,Lee2011Geo}. 
The authors of~\cite{tal2020accurate} demonstrated accurate tracking of aggressive quadrotor trajectories up to \SI{12.9}{\meter\per\second} using a cascaded geometric controller with Incremental Nonlinear Dynamic Inversion (INDI). 
In~\cite{torrente2021datadriven}, NMPC leveraging data driven methods to improve model fidelity was used to achieve state of the art tracking performance at speeds up to \SI{14}{\meter\per\second}. 
Control commands were calculated off-board the quadrotor and sent via wireless communication due to the increased computational overhead from the learned model. 
Because the model parameters are obtained offline, it can not adapt to online parametric changes such as payloads or a reduction in actuator efficacy. 

As the authors in~\cite{torrente2021datadriven} correctly point out, INDI and adaptive control approaches coupled with traditional geometric controllers are purely reactive and have no ability to plan over a prediction horizon. 
To address this, we couple an $\mathcal{L}_1$ adaptive control law with a nonlinear MPC, therefore taking advantage of the prediction horizon and fully exploiting the nonlinearities of the system for maximum performance.
This selection is further motivated by the recent comparative study between geometric and nonlinear MPC methods for agile quadrotor control in \cite{sun2021comparative}
The controller is evaluated against several state of the art controllers and a wide variety of test conditions both in simulation and reality.

\section{Methodology}\label{sec:method}
\subsection{Notation}
We define the World $W$ and Body $B$ frames with orthonormal basis i.e. $\{\bm{x}_W, \bm{y}_W, \bm{z}_W\}$.
The frame $B$ is located at the center of mass of the quadrotor.
All four rotors are assumed to  be located on the $xy$-plane of frame $B$, as depicted in Fig.~\ref{fig:quad_top_down_diagram}.
A vector from coordinate $\bm{p}_1$ to $\bm{p}_2$ expressed in the $W$ frame is written as: $_W\bm{v}_{12}$.
If the vector's origin coincide with the frame it is described in, we drop the frame index, e.g. the quadrotor position is denoted as $\bm{p}_{WB}$.
Furthermore, we use unit quaternions $\bm{q} = (q_w, q_x, q_y, q_z)$ with $\|\bm{q}\| = 1$ to represent orientations, such as the attitude state of the quadrotor body $\bm{q}_{WB}$.
Finally, full SE3 transformations, such as changing the frame of reference from body to world for a point $\bm{p}_{B1}$, can be described by $_W\bm{p}_{B1} = _W\bm{t}_{WB} + \bm{q}_{WB} \odot \bm{p}_{B1}$.
Note the quaternion-vector product denoted by $\odot$ representing a rotation of the vector by the quaternion as in $\bm{q} \odot \bm{v} = \bm{q} \cdot [0, \bm{v}^\intercal]^\intercal \cdot \bar{\bm{q}}$, where $\bar{\bm{q}}$ is the quaternion's conjugate.

\subsection{Quadrotor Vehicle Dynamics}
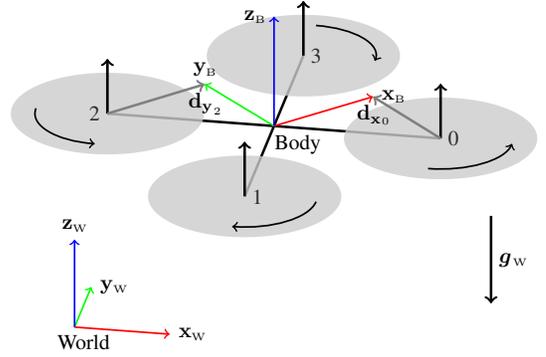
\begin{figure}[t]
\centerline{
\resizebox{7.8cm}{!}{
    \tdplotsetmaincoords{65}{10}
\begin{tikzpicture}[tdplot_main_coords, scale=2]

\draw[very thick] (-1.4,0,1) -- (1.4,0,1);
\draw[very thick] (0,-1.4,1) -- (0,1.4,1);

\draw[draw=none, fill=gray!40, opacity=0.8] (0,1.4,1) circle (0.8) node[right] {3};
\draw[draw=none, fill=gray!40, opacity=0.8] (1.4,0,1) circle (0.8) node[right] {0};
\draw[draw=none, fill=gray!40, opacity=0.8] (0,-1.4,1) circle (0.8) node[right] {1};
\draw[draw=none, fill=gray!40, opacity=0.8] (-1.4,0,1) circle (0.8) node[left] {2};

\draw[very thick,->] (0,1.4,1) -- (0,1.4,1.5);
\draw[very thick,->] (1.4,0,1) -- (1.4,0,1.5);
\draw[very thick,->] (0,-1.4,1) -- (0,-1.4,1.5);
\draw[very thick,->] (-1.4,0,1) -- (-1.4,0,1.5);	

\draw[thick,->] (0,2.0,1) arc (90:0:0.6);
\draw[thick,<-] (2.0,0,1) arc (0:-90:0.6);
\draw[thick,->] (0.6,-1.4,1) arc (0:-90:0.6);
\draw[thick,->] (-2.0,0,1) arc (180:270:0.6);

\draw[thick,->,color=red,text=black] (0,0,1) -- (0.7071,0.7071,1) node[right] {$\vect{x}{}{\bfr}$};
\draw[thick,->,color=green,text=black] (0,0,1) -- (-0.7071,0.7071,1) node[above] {$\vect{y}{}{\bfr}$};	
\draw[thick,->,color=blue,text=black] (0,0,1) -- (0,0,2) node[left] {$\vect{z}{}{\bfr}$};		
\node[draw=none] at (0.2,0,0.85) {Body};	

\draw[very thick,->,color=gray,text=black] (1.4,0,1) --  (0.7071,0.7071,1) node[below] {$\vect{d_x}{}{0}$};
\draw[very thick,->,color=gray,text=black] (-1.4,0,1) --  (-0.7071,0.7071,1) node[below] {$\vect{d_y}{}{2}$};

\draw[thick,->,color=red,text=black] (-1.5,-1,-0.5) -- (-0.7,-1,-0.5) node[right] {$\vect{x}{}{\wfr}$};		
\draw[thick,->,color=green,text=black] (-1.5,-1,-0.5) -- (-1.5,-0.2,-0.5) node[right] {$\vect{y}{}{\wfr}$};		
\draw[thick,->,color=blue,text=black] (-1.5,-1,-0.5) -- (-1.5,-1,0.3) node[above] {$\vect{z}{}{\wfr}$}; %
\node[draw=none] at (-1.6,0,-1.1) {World};

\draw[very thick,->,color=black,text=black] (2,-1.0,0.8) -- node[right] {$\bm{g}\vect{}{}{\wfr}$} (2,-1.0,0);

\end{tikzpicture}
}}
\caption{Diagram of the quadrotor model with the world and body frames and propeller numbering convention.}
\label{fig:quad_top_down_diagram}
\end{figure}

The quadrotor system dynamics are given by
    \begin{align}
        \dot{x} =
        \begin{bmatrix}
            \dot{\bm{p}}_{WB} \\
            \dot{\bm{q}}_{WB} \\
            \dot{\bm{v}}_{WB} \\
            \dot{\bm{\omega}}_B
        \end{bmatrix}
        = 
        \begin{bmatrix}
            \bm{v}_W \\
            \bm{q}_{WB} 
            \begin{bmatrix}
                0 &
                \bm{\omega}^\intercal_B/2
            \end{bmatrix}^\intercal
            \\
            \bm{q}_{WB}\odot \bm{T}_B/m + \bm{g} \\
            \bm{J}^{-1}[\bm{\tau}_B-\bm{\omega}_B \times J\bm{\omega}_B]
        \end{bmatrix} \; ,
    \end{align}
where $\bm{g} = [0, 0, \SI{-9.81}{\meter\per\second^2}]^T$ denotes Earth's gravity, $T_B$ is the collective thrust from the 4 rotors, $\bm{J} = \text{diag}(J_x, J_y, J_z)$ is the diagonal moment of inertia matrix, $m$ is the quadrotor mass, and $\bm{\tau}_B$ is the body torque:
\begin{align}
    \bm{T}_B &=
    \begin{bmatrix} 0 \\ 0 \\ \sum T_i
    \end{bmatrix}
    &
    \!\!\text{and}\!\! \quad 
    \bm{\tau}_B &= \! \bm{P} \begin{bmatrix}
        T_0\\T_1\\T_2\\T_3
    \end{bmatrix}
\end{align}
    Where $\bm{P}$ is the thrust allocation matrix given by
\begin{align}
    \bm{P} = \begin{bmatrix}
        -d_{x_0} & -d_{x_1} & d_{x_2} & d_{x_3} \\
        d_{y_0} & - d_{y_1} & - d_{y_2} & d_{y_3} \\
        -c_\tau & c_\tau & -c_\tau & c_\tau
    \end{bmatrix}
\end{align}
where $d_{x_i}$, $d_{y_i}$, $c_\tau$ for $i={0,1,2,3}$ are the distances from each rotor, $i$, to the respective Body frame axis and the rotor drag torque constant respectively following the rotor positions and spin configurations denoted in Figure \ref{fig:quad_top_down_diagram} and obeying the right hand rule.
For discrete time, an explicit Runge-Kutta method of 4th order is used:
\begin{align}
    \bm{x}_{k+1} = f_{RK4}(\bm{x}_k,\bm{u}_k,\delta t).
\end{align}
    
 \subsection{MPC Formulation}
 We construct a quadratic optimization problem using a multi-shooting scheme and solve the following discretized nonlinear optimal control problem
 \begin{gather}
     \min\limits_{\bm{u}} \bm{x}_N^T \bm{Q} \bm{x}_N + \sum_{k=0}^{N-1} \bm{x}_k^T \bm{Q} \bm{x}_k + \bm{u}_k^T \bm{R} \bm{u}_k \\
     \st \quad \bm{x}_{k+1} = \bm{f}_{RK4}(\bm{x}_k, \bm{u}_k, \delta t)\nonumber\\
     \begin{aligned}
     \bm{x}_0 &= \bm{x}_{init}\nonumber &
     \bm{u}_{min} &\leq \bm{u}_k \leq \bm{u}_{max}\nonumber
     \end{aligned}
 \end{gather}
 as a sequential quadratic program (SQP) executed in a real-time iteration scheme~\cite{Diehl2006springer}.
 We discretize the system evolution into $N$ steps over a time horizon $T$ and constrain the input to be between $\bm{0} \leq \bm{u}_k \leq \bm{u}_{max} $.
 The optimal control problem is implemented using the open source ACADO toolkit~\cite{Houska2011acado}.
    
\subsection[L1-Adaptive Augmentation]{$\mathcal{L}_1$-Adaptive Augmentation}
We implement the $\mathcal{L}_1$ adaptive controller using a nonlinear reference model~\cite{wang2012l1}, which estimates both matched and unmatched uncertainties using a piecewise constant adaptation law~\cite{li20121L1MIMOPiecewise, xargay2010L1MIMOUnmatched}.
The derivation is similar to~\cite{pravitra2020l1adaptive}, however we account for the uncertainties directly at the rotor thrust level.
First, define
$\bm{R}_B^I = \begin{bmatrix}
    \bm{e}^B_x, \bm{e}^B_y,\bm{e}^B_z
\end{bmatrix}$
as the rotation matrix from the body frame to the inertial frame. We can then rewrite the dynamics to account for both matched and unmatched uncertainties as
\begin{align}
    \dot{\bm{v}}_{WB} &= \frac{\bm{T}_B}{m} \bm{e}^B_z + \bm{g} + \frac{\bm{T}_B}{m} \bm{R}_B^I \bm{\varsigma} - \bm{C_d} \bm{v}_{WB},\\
    \dot{\bm{\omega}}_{B} &= \bm{J}^{-1}[\bm{\tau}_B-\bm{\omega}_B \times \bm{J}\bm{\omega}_B + \bm{\xi}],
\end{align}
where
$\bm{\varsigma} = \begin{bmatrix}
    \varsigma_x, \varsigma_y, \varsigma_z
\end{bmatrix}^T$
is the uncertainty appearing in the linear accelerations,
$\bm{\xi} = \begin{bmatrix}
    \xi_x, \xi_y, \xi_z
\end{bmatrix}^T$
is the uncertainty appearing in the angular accelerations, and $\bm{C_d}$ is a matrix of linear drag coefficients.
Since a quadrotor is an underactuated system, capable of providing linear acceleration only along its body z-axis, the unmatched uncertainties defined as ${\bm{\sigma}_{um} = \begin{bmatrix}
    \varsigma_x, \varsigma_y
\end{bmatrix}^T}$,
appear purely in the X and Y linear accelerations.
This can be thought of as existing in the null space of the controllability matrix and therefore cannot be compensated for directly.
Then, what remains are the matched uncertainties
$\bm{\sigma}_m = \begin{bmatrix}
    \varsigma_z, \xi_x, \xi_y, \xi_z
\end{bmatrix}^T$
which can be compensated for directly.

Next, consider the reduced state variable,
${\bm{z} = \begin{bmatrix}
\bm{v}_{WB}, \bm{\omega}_B 
\end{bmatrix}}$.
Its derivative can be broken up as a function of the nominal and uncertain dynamic behavior as follows:
\begin{align}
    \dot{\bm{z}} = \bm{f}(\bm{R}_B^I) + \bm{g}(\bm{R}_B^I)(\bm{u}_{\mathcal{L}1} +\bm{\sigma}_m) + \bm{g}^\perp(\bm{R}_B^I) \bm{\sigma}_{um} \; ,
\end{align}
where $\bm{f}(\bm{R}_B^I)$ is the desired dynamics defined as
\begin{align}
    \bm{f}(\bm{R}_B^I) = 
    \begin{bmatrix}
        \bm{g} + \frac{\bm{T}_{MPC}}{m} \bm{e}^B_z - \bm{C_d}\bm{v}_{WB}\\
        \bm{J}^{-1} \bm{\tau}_B - \bm{\omega}_B \times \bm{J}\bm{\omega}_B]
    \end{bmatrix} \; .
\end{align}
The single rotor thrusts in expression of $\bm{\tau}_B$ are obtained from the solution of the NMPC at the current time.

Define $\bm{g}(\bm{R}_B^I)$ as the uncertainty in the matched component of the dynamics, and $\bm{g}^{\perp}(\bm{R}_B^I)$ the unmatched component: 
\begin{align}
    \bm{g}(\bm{R}_B^I) &= 
    \begin{bmatrix}
        \frac{\bm{e}^B_z}{m}, \frac{\bm{e}^B_z}{m}, \frac{\bm{e}^B_z}{m}, \frac{\bm{e}^B_z}{m}\\
        \bm{J}^{-1} \bm{P}
    \end{bmatrix} &
    \bm{g}^{\perp}(\bm{R}_B^I) &= 
    \begin{bmatrix}
        \frac{\bm{e}^B_x}{m}, \frac{\bm{e}^B_y}{m}\\
        \bm{0}_{3\times1}, \bm{0}_{3\times1}
    \end{bmatrix} \; .
\end{align}
Let $\bm{u}_{\mathcal{L}_1}$ be the adaptive control input which can act as a standalone controller, or complement the NMPC signal via addition.
Define the $\mathcal{L}_1$ observer as 
\begin{align}
    \dot{\hat{\bm{z}}} = \bm{f}(\bm{R}_B^I) + \bm{g}(\bm{R}_B^I)(\bm{u}_{\mathcal{L}_1} + \hat{\bm{\sigma}}_m) + \bm{g}^{\perp}(\bm{R}_B^I) \hat{\bm{\sigma}}_{um} + \bm{A}_s \Tilde{\bm{z}} \; ,
    \nonumber
\end{align}
where $\Tilde{\bm{z}} = \hat{\bm{z}} - \bm{z}$ and noting that $\bm{z}$ is the state obtained from an estimator, and $\hat{\bm{z}}$ is the state predicted from the $\mathcal{L}_1$ observer.
Define $\bm{\Phi} = \bm{A}_s^{-1}(e^{(\bm{A}_s T_s)}-\bm{I})$ where $T_s$ is the time step  and $\bm{A}_s$ is a Hurwitz matrix which represents the adaptation gains.
Then the piecewise-constant adaptation law is given by
    \begin{align}
        \begin{bmatrix}
            \hat{\bm{\sigma}}_m(i T_s) \\
            \hat{\bm{\sigma}}_{um}(i T_s)
        \end{bmatrix}
        = -\bm{I}_{6\times6} \bm{G}^{-1}(i T_s) \bm{\Phi}^{-1} \bm{\mu}(i T_s) \; ,
    \end{align}
where $\bm{G}(iT_s) = [\bm{g}(\bm{R}_B^I), \bm{g}^{\perp}(\bm{R}_B^I)]$ and $\bm{\mu} = e^{\bm{A}_s T_s}\Tilde{\bm{z}}(i T_s)$  are evaluated at time step $i$. 
Next, define a first order, strictly proper continuous time filter $\bm{C}(s)$.
The $\mathcal{L}_1$ control law is then
    \begin{align}
        \bm{u}_{\mathcal{L}1} = -\bm{C}(s)\hat{\bm{\sigma}}_m \; .
    \end{align}
In practice, we implement the control law in discrete time as
\begin{align}
    \bm{u}_{\mathcal{L}1,k} = \bm{u}_{\mathcal{L}1,k-1} e^{-\bm{\omega}_{co} T_s} - \hat{\bm{\sigma}}_{m,k} (1 - e^{-\bm{\omega}_{co} T_s}) \; ,
\end{align}
where $\bm{\omega}_{co}$ is the cutoff frequency of the strictly proper first order filter. 
Finally, the discrete time $\mathcal{L}_1$ observer can be propagated forward in time via
\begin{align}
    \small
    \hat{\bm{z}}_{k+1} = \hat{\bm{z}}_k + [\bm{f}_k + \bm{g}_k(\bm{u}_{\mathcal{L}1,k} + \hat{\bm{\sigma}}_{m,k}) + \bm{g}_k^{\perp}\hat{\bm{\sigma}}_{um,k} + \bm{A}_s\Tilde{\bm{z}}_k] T_s \; .
    \nonumber
\end{align}

\section{Experiments and Results}\label{sec:experiments}
Our experiments are designed to answer the following research questions: 
(i) How does our proposed $\mathcal{L}_1$-NMPC compare to data-driven MPC methods?
(ii) To what extent can $\mathcal{L}_1$-NMPC
react to both parametric and non-parametric disturbances in real world tests? 
(iii) Does the proposed $\mathcal{L}_1$-NMPC
generalize across test scenarios without the need for gain tuning? 
We set out to answer these questions through a variety of simulation and real world tests which cover large external disturbances and agile maneuvers. 
The experiments start out with a set of different model predictive controllers which are gradually eliminated over the course of the experiments based on their tracking performance and computational complexity. 
We point the reader to our corresponding video for a better understanding of the level of flexibility and robustness our proposed architecture can demonstrate.

\subsection{Simulation}
We begin by implementing several state of the art controllers in simulation including a Single Rotor Thrust NMPC (SRT-NMPC), a data-driven NMPC (GP-MPC) from~\cite{torrente2021datadriven}, MPPI with Baseline Control from~\cite{pravitra2020l1adaptive}, INDI-NMPC from~\cite{sun2021comparative}, and our proposed $\mathcal{L}_1$-NMPC with and without an aerodynamic model included in the underlying NMPC. 
All baseline controllers have an aerodynamic model enabled by default. 
We use the open source Gazebo simulator~\cite{koenig2004design} with the the AscTec Hummingbird quadrotor model using the RotorS extension~\cite{Furrer2016_RotorS}.
Performance in simulation is measured by comparing positional tracking errors on a set of circular reference trajectories. 
These trajectories feature a radius of \SI{5}{\meter} and vary in peak velocities from 2.5-10\si{\meter\per\second}.

\subsubsection{Without Disturbance}
First, we fly these trajectories with a model that best represents our knowledge of the dynamics within the RotorS simulation. 
In these cases, the mass, inertia, drag, and rotor arm lengths of the simulated quadrotor are perfectly known.
The model parameters used in the NMPC are identical to the parameters in the RotorS simulator.
These experiments without disturbance serve to understand the maximum achievable performance by each method in case of perfect model identification.

\begin{figure}[t]
    \centering
    \vspace{0.5em}
    \definecolor{mycolor1}{rgb}{0.00000,1.00000,1.00000}%
\begin{tikzpicture}

\begin{axis}[%
width=2.8521in,
height=1.866in,
at={(0.0758in,0.481in)},
scale only axis,
xmin=2,
xmax=10.5,
xlabel style={font=\color{white!15!black}},
xlabel={Velocity, m/s},
label style={font=\small},
tick label style={font=\scriptsize},
ymode=log,
ymin=0.005,
ymax=2,
yminorticks=true,
ylabel style={font=\color{white!15!black}},
ylabel={RMSE [m]},
axis background/.style={fill=white},
xmajorgrids,
ymajorgrids,
yminorgrids,
legend style={at={(0.03,0.97)}, 
  fill opacity=0.8,
  draw opacity=1,
  text opacity=1,anchor=north west, legend cell align=left, align=left, draw=white!15!black}
]
\addplot [color=mycolor1, mark=o, mark options={solid, mycolor1}]
  table[row sep=crcr]{%
2.5	0.017\\
4	0.022\\
6	0.031\\
8	0.04\\
10	0.053\\
};
\addlegendentry{\scriptsize GP-MPC}

\addplot [color=blue, mark=o, mark options={solid, blue}]
  table[row sep=crcr]{%
2.5	0.016\\
4	0.034\\
6	0.037\\
8	0.054\\
10	0.062\\
};
\addlegendentry{\scriptsize Single Rotor NMPC}

\addplot [color=green, mark=o, mark options={solid, green}]
  table[row sep=crcr]{%
2.5	0.012\\
4	0.016\\
6	0.032\\
8	0.048\\
10	0.06\\
};
\addlegendentry{\scriptsize INDI-NMPC}

\addplot [color=black, mark=o, mark options={solid, black}]
  table[row sep=crcr]{%
2.5	0.056\\
4	0.128\\
6	0.272\\
8	0.462\\
10	1.607\\
};
\addlegendentry{\scriptsize MPPI}

\addplot [color=red, dashdotted, mark=o, mark options={solid, red}]
  table[row sep=crcr]{%
2.5	0.007\\
4	0.014\\
6	0.03\\
8	0.052\\
10	0.079\\
};
\addlegendentry{\scriptsize $\mathcal{L}_1$-NMPC, No Aero}

\addplot [color=red, mark=o, mark options={solid, red}]
  table[row sep=crcr]{%
2.5	0.006\\
4	0.012\\
6	0.025\\
8	0.041\\
10	0.054\\
};
\addlegendentry{\scriptsize $\mathcal{L}_1$-NMPC, Aero}

\end{axis}
\end{tikzpicture}%
    \vspace{-8pt}
    \caption{Simulated tracking accuracy of increasing speed, on a \SI{5}{\meter} radius circle trajectory in the RotorS simulation environment. Tracking accuracy is given by RMSE from the reference trajectory. Each point represents the accumulated positional tracking error across the trajectory for a given peak velocity.}
    \label{fig:RotorSCircle_error}
\end{figure}
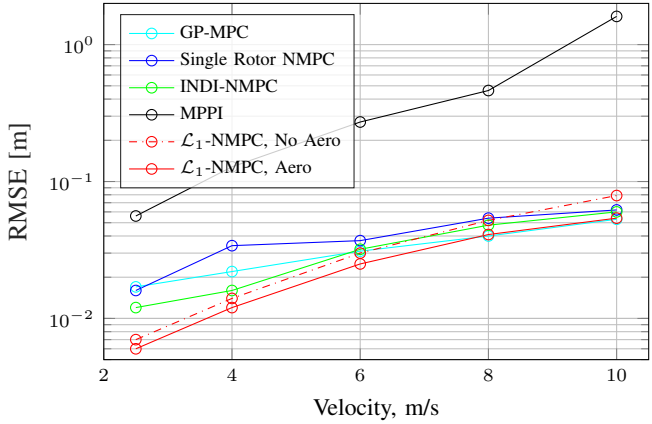

We include the timings for each of the tested controllers in Table~\ref{tab:Timing} and show the tracking performance on a semi-log scale for the increasing speed circle trajectories in Figure~\ref{fig:RotorSCircle_error}.
Our $\mathcal{L}_1$-NMPC \emph{without} an aerodynamic model  outperforms the state of the art INDI-NMPC and GP-MPC in all cases except for the fastest circle trajectories, indicating that the adaptation law is able to compensate partially for unmodeled aerodynamics. 

Once we enable the drag model, the $\mathcal{L}_1$-NMPC outperforms all proposed methods, however we note that the performance improvement is less than \SI{1}{\centi\meter} RMSE over the course of a \SI{60}{\second} long trajectory. 
We expect the performance benefit to be small since the model parameters perfectly match those described in the simulator. 
In contrast to all other approaches, MPPI performed an order of magnitude worse.
We therefore do not consider it as a viable candidate moving forward.
SRT-NMPC with a linear drag model performs similarly to GP-MPC, but with 80\% less computational overhead and therefore forms the baseline for the remaining simulation trials.

\begin{table}[t!]
    \centering
\vspace{0.6em}
\resizebox{\columnwidth}{!}{%
\begin{tabular}{c|c|c|c|c|c|}
\cline{2-6}
  & \multicolumn{5}{c|}{\textbf{Model}} \\ \cline{2-6}
\textbf{} & \textbf{GP-MPC} & \textbf{MPPI} & \textbf{NMPC} & \textbf{INDI-NMPC} & \textbf{L1-NMPC} \\ \hline
\multicolumn{1}{|c|}{\begin{tabular}[c]{@{}c@{}}\textbf{Avg. dt}\\ $[\si{\milli\second}]$ \end{tabular}} & 4.13 & 23.13 & 0.81 & 0.82 & 0.82 \\ \cline{1-6}

\end{tabular}
}
    \caption{Average controller update times running on an Intel Core i7-8750H CPU @ 2.20GHz laptop with 16 Gb of RAM and Nvidia GeForce GTX 1060 using CUDA 11.2}
    \label{tab:Timing}

\end{table}

\subsubsection{Model Mismatch}
Next, we inject three different forms of model mismatch into the model used for the NMPC. 
The same increasing speed circle trajectories from the previous section are used in this analysis. 
First, the mass of the quadrotor is increased by \SI{660}{\gram}, representing a 90\% increase from the nominal mass of the system.
Second, the inertias on all axis are doubled in the simulator. 
Finally, the right side rotor arm lengths are reduced by 25\%, representing a persistent center of gravity offset.

Since the baseline NMPC does not contain any integrator action, we expect any pure-mass disturbances to result in a steady-state offset in target Z-height.
In an attempt to make a fair comparison, we add an additional comparison case which embeds an integrator state on positional error into the NMPC. 
The gains for all controllers are not adjusted from the nominal tests in the previous section. 
Table~\ref{tab:RotorSCircleError} provides the results for these parametric disturbance cases. 
\begin{table}[t!]

\centering

\resizebox{\columnwidth}{!}{%
\begin{tabular}{cc|c|cc|cc|cc|}
\cline{3-9}
 &  & \multicolumn{7}{c|}{\textbf{Model}} \\ \cline{3-9} 
\textbf{} & \textbf{} & \textbf{NMPC} & \multicolumn{2}{c|}{\textbf{NMPC+I}} & \multicolumn{2}{c|}{\textbf{INDI-NMPC}} & \multicolumn{2}{c|}{\textbf{L1-NMPC}} \\ \hline
\multicolumn{1}{|c|}{\textbf{Dist.}} & \begin{tabular}[c]{@{}c@{}}$\mathbf{v_{peak}}$\\ $[\si{\meter\per\second}]$\end{tabular} & \begin{tabular}[c]{@{}c@{}}RMSE \\ $[\si{\meter}]$\end{tabular} & \begin{tabular}[c]{@{}c@{}}RMSE\\ $[\si{\meter}]$\end{tabular} & \cellcolor[HTML]{EFEFEF}\textbf{\%}$\downarrow$ & \begin{tabular}[c]{@{}c@{}}RMSE\\ $[\si{\meter}]$\end{tabular} & \cellcolor[HTML]{EFEFEF}\textbf{\%}$\downarrow$ & \begin{tabular}[c]{@{}c@{}}RMSE\\ $[\si{\meter}]$\end{tabular} & \cellcolor[HTML]{EFEFEF}\textbf{\%}$\downarrow$ \\ \hline
\multicolumn{1}{|c|}{} & 2.5 & 0.434 & 0.029 & \cellcolor[HTML]{EFEFEF}93 & 0.434 & \cellcolor[HTML]{EFEFEF}0 & 0.007 & \cellcolor[HTML]{EFEFEF}\textbf{98} \\
\multicolumn{1}{|c|}{} & 4 & 0.441 & 0.045 & \cellcolor[HTML]{EFEFEF}90 & 0.440 & \cellcolor[HTML]{EFEFEF}0 & 0.010 & \cellcolor[HTML]{EFEFEF}\textbf{98} \\
\multicolumn{1}{|c|}{} & 6 & 0.471 & 0.091 & \cellcolor[HTML]{EFEFEF}81 & 0.470 & \cellcolor[HTML]{EFEFEF}0 & 0.020 & \cellcolor[HTML]{EFEFEF}\textbf{96} \\
\multicolumn{1}{|c|}{} & 8 & 0.532 & 0.167 & \cellcolor[HTML]{EFEFEF}69 & 0.537 & \cellcolor[HTML]{EFEFEF}0 & 0.032 & \cellcolor[HTML]{EFEFEF}\textbf{94} \\
\multicolumn{1}{|c|}{\multirow{-5}{*}{\rotatebox[origin=c]{90}{\textbf{Mass}}}} & 10 & \emph{Crash} & \emph{Crash} & \cellcolor[HTML]{EFEFEF}\emph{?} & \emph{Crash} & \cellcolor[HTML]{EFEFEF}\emph{?} & \emph{Crash} &  \cellcolor[HTML]{EFEFEF}\emph{?}\\ \hline
\multicolumn{1}{|c|}{} & 2.5 & 0.02 & 0.033 & \cellcolor[HTML]{EFEFEF}-65 & 0.009 & \cellcolor[HTML]{EFEFEF}55 & 0.007 & \cellcolor[HTML]{EFEFEF}\textbf{65} \\
\multicolumn{1}{|c|}{} & 4 & 0.016 & 0.043 & \cellcolor[HTML]{EFEFEF}-169 & 0.015 & \cellcolor[HTML]{EFEFEF}6 & 0.014 & \cellcolor[HTML]{EFEFEF}\textbf{13} \\
\multicolumn{1}{|c|}{} & 6 & 0.037 & 0.080 & \cellcolor[HTML]{EFEFEF}-116 & 0.028 & \cellcolor[HTML]{EFEFEF}\textbf{24} & 0.030 & \cellcolor[HTML]{EFEFEF}19 \\
\multicolumn{1}{|c|}{} & 8 & 0.070 & 0.150 & \cellcolor[HTML]{EFEFEF}-114 & 0.048 & \cellcolor[HTML]{EFEFEF}\textbf{31} & 0.052 & \cellcolor[HTML]{EFEFEF}26 \\
\multicolumn{1}{|c|}{\multirow{-5}{*}{\rotatebox[origin=c]{90}{\textbf{Inertia}}}} & 10 & 0.081 & 0.240 & \cellcolor[HTML]{EFEFEF}-196 & .074 & \cellcolor[HTML]{EFEFEF}\textbf{9} &  0.079 & \cellcolor[HTML]{EFEFEF}2 \\ \hline
\multicolumn{1}{|c|}{} & 2.5 & 0.078 & 0.056 & \cellcolor[HTML]{EFEFEF}28 & 0.008 & \cellcolor[HTML]{EFEFEF}90 & 0.007 & \cellcolor[HTML]{EFEFEF}\textbf{91}  \\
\multicolumn{1}{|c|}{} & 4 & 0.049 & 0.038 & \cellcolor[HTML]{EFEFEF}22 & 0.015 & \cellcolor[HTML]{EFEFEF}\textbf{69} & 0.016 & \cellcolor[HTML]{EFEFEF}67 \\
\multicolumn{1}{|c|}{} & 6 & 0.052 & 0.088 & \cellcolor[HTML]{EFEFEF}-69 & 0.032 & \cellcolor[HTML]{EFEFEF}\textbf{38} & 0.033 & \cellcolor[HTML]{EFEFEF}37 \\
\multicolumn{1}{|c|}{} & 8 & 0.057 & 0.151 & \cellcolor[HTML]{EFEFEF}-165 & 0.053 & \cellcolor[HTML]{EFEFEF}\textbf{7} & 0.056 & \cellcolor[HTML]{EFEFEF}2 \\
\multicolumn{1}{|c|}{\multirow{-5}{*}{\rotatebox[origin=c]{90}{\textbf{Arm Length}}}} & 10 & \textbf{0.083} & 0.241 & \cellcolor[HTML]{EFEFEF}-190 & 0.083 & \cellcolor[HTML]{EFEFEF}0 & 0.087 &  \cellcolor[HTML]{EFEFEF}-5\\ \hline
\end{tabular}

}
\caption{Tracking performance of increasing speed, \SI{5}{\meter} radius circle trajectories with mass, inertia, and rotor arm length disturbances in the RotorS simulator. Each table entry represents the tracking RMSE over the course of a single trajectory for different speeds.
In the Mass cases, we increase the mass of the quadrotor by \SI{660}{\gram} representing a 90\% increase in mass. Similarly, in the Inertia cases, we double the inertia of the quadrotor. Finally, in the Arm Length cases, we increase the length of the right side rotor arms by 25\% representing a significant shift to the center of gravity.}
\label{tab:RotorSCircleError}

\end{table}

As parametric disturbances are introduced into the test scenarios, the proposed $\mathcal{L}_1$-NMPC architecture demonstrates a sizeable advantage over the other state of the art methods.
Only the NMPC with integrator action and $\mathcal{L}_1$-NMPC show any ability to compensate for mass mismatch, with $\mathcal{L}_1$-NMPC reducing the tracking error by over 90\% in all cases.
We note that all controllers fail to complete the \SI{10}{\meter\per\second} circle trajectory with a \SI{660}{\gram} payload. 
The thrust requirements to fly this case exceed what is available on the simulated quadrotor which leads to the NMPC solution failing to converge in all cases.
In the inertia and rotor arm length test cases, INDI-NMPC and $\mathcal{L}_1$-NMPC perform almost identically.
The performance delta between the two methods is less than \SI{5}{\mm} across these 10 cases.
Our adaptive architecture demonstrates robustness to these uncertainties in addition to providing a performance benefit when the model is well known.
We are able to immediately identify various forms of disturbance and swiftly reject it without any gain tuning or model learning.

\subsection{Real World Experiments}
We test our proposed $\mathcal{L}_1$-NMPC controller performance on a quadrotor with mass \SI{750}{\gram} outfitted with a Jetson TX2 on-board computer and Radix flight controller with our own custom low-level flight control firmware.
The flight controller accepts single rotor thrust inputs and performs closed loop rotor speed control. 
The quadrotor has a thrust to weight ratio of about 4.5.
We run all controllers completely onboard the quadrotor and solve the optimal control problem at \SI{100}{\hertz}.
A Vicon motion capture system\footnote{\href{Vicon}{https://www.vicon.com/}} provides pose updates at \SI{400}{\hertz} which are fused with the Inertial Measurement Unit (IMU) via an Extended Kalman Filter (EKF) for state estimation. 

To demonstrate our approach can significantly improve tracking performance under model mismatch and aerodynamic disturbance, we conduct experiments in five settings:
\begin{itemize}
    \item Setting (i): Increasing speed \SI{5}{\meter} radius circle trajectories up to \SI{10}{\meter\per\second} to compare performance to GP-MPC and INDI-NMPC in a nominal setting.
    \item Setting (ii): \SI{2}{\meter\per\second} circle trajectory with a \SI{450}{\gram} unknown payload to show tracking performance when a large mass mismatch is present. 
    \item Setting (iii): \SI{2}{\meter\per\second} circle with external aerodynamic forces to show rapid disturbance rejection.
    \item Setting (iv): Mildly aggressive flight with a top speed of \SI{11.9}{\meter\per\second} with a \SI{100}{\gram} slung payload to show agile flight is possible with unknown payloads.
    \item Setting (v): Highly aggressive flight with a top speed of \SI{19.4}{\meter\per\second}  without payload to demonstrate capability of accurate tracking near the system limits.
\end{itemize}

In Setting (i), we perform the same increasing speed circle trajectories from the simulation experiments and compare the tracking performance of our method to GP-MPC, SRT-NMPC, and INDI-NMPC.
Table~\ref{tab:FlightArenaIncreasingSpeedCircleTable} shows the
results of this comparison.
As can be seen, even without an aerodynamic model, 
our $\mathcal{L}_1$-NMPC architecture outperforms GP-MPC by over 70\% and has a slight advantage over INDI-NMPC, indicating that the adaptive controller is compensating for both model mismatch and aerodynamic disturbances.
Additionally, we show that SRT-NMPC with a linear aerodynamic model matches the performance of GP-MPC and therefore use SRT-NMPC as the baseline moving forward due to its computational advantage over GP-MPC.

\begin{table}[t!]
    \centering
\vspace{0.7em}
\resizebox{\columnwidth}{!}{%
\begin{tabular}{c|c|cc|cc|cc|cc|}
\cline{2-10}
  & \multicolumn{9}{c|}{\textbf{Model}} \\ \cline{2-10}
\textbf{} & \textbf{GP-MPC} & \multicolumn{2}{c|}{\textbf{SRT-NMPC}} & \multicolumn{2}{c|}{\textbf{INDI-NMPC}} & \multicolumn{2}{c|}{\textbf{L1-NMPC No Aero}} & \multicolumn{2}{c|}{\textbf{L1-NMPC w/ Aero}}\\ \hline
\multicolumn{1}{|c|}{\begin{tabular}[c]{@{}c@{}}$\mathbf{v_{peak}}$\\ $[\si{\meter\per\second}]$\end{tabular}} & \begin{tabular}[c]{@{}c@{}}RMSE \\ $[\si{\meter}]$\end{tabular} & \begin{tabular}[c]{@{}c@{}}RMSE\\ $[\si{\meter}]$\end{tabular} & \cellcolor[HTML]{EFEFEF}\textbf{\%}$\downarrow$ & \begin{tabular}[c]{@{}c@{}}RMSE\\ $[\si{\meter}]$\end{tabular} & \cellcolor[HTML]{EFEFEF}\textbf{\%}$\downarrow$ & \begin{tabular}[c]{@{}c@{}}RMSE\\ $[\si{\meter}]$\end{tabular} & \cellcolor[HTML]{EFEFEF}\textbf{\%}$\downarrow$ & \begin{tabular}[c]{@{}c@{}}RMSE\\ $[\si{\meter}]$\end{tabular} & \cellcolor[HTML]{EFEFEF}\textbf{\%}$\downarrow$ \\ \hline
\multicolumn{1}{|c|}{2.5} & 0.109 & 0.070 & \cellcolor[HTML]{EFEFEF}36 & 0.038 & \cellcolor[HTML]{EFEFEF}65 & 0.022 & \cellcolor[HTML]{EFEFEF}80 & 0.016 & \cellcolor[HTML]{EFEFEF}\textbf{85} \\
\multicolumn{1}{|c|}{4} & 0.103 & 0.086 & \cellcolor[HTML]{EFEFEF}17 & 0.052 & \cellcolor[HTML]{EFEFEF}50 & 0.026 & \cellcolor[HTML]{EFEFEF}75 & 0.021 & \cellcolor[HTML]{EFEFEF}\textbf{80} \\
\multicolumn{1}{|c|}{6} & 0.129 & 0.103 & \cellcolor[HTML]{EFEFEF}20 & 0.055 & \cellcolor[HTML]{EFEFEF}57 & 0.035 & \cellcolor[HTML]{EFEFEF}73 & 0.034 & \cellcolor[HTML]{EFEFEF}\textbf{74} \\
\multicolumn{1}{|c|}{8} & 0.154 & 0.153 & \cellcolor[HTML]{EFEFEF}1 & 0.058 & \cellcolor[HTML]{EFEFEF}62 & 0.057 & \cellcolor[HTML]{EFEFEF}63 & 0.041 & \cellcolor[HTML]{EFEFEF}\textbf{73} \\
\multicolumn{1}{|c|}{10} & 0.203 & 0.223 & \cellcolor[HTML]{EFEFEF}-10 & 0.048 & \cellcolor[HTML]{EFEFEF}76 & 0.088 &  \cellcolor[HTML]{EFEFEF}57 & 0.047 & \cellcolor[HTML]{EFEFEF}\textbf{77}\\ \hline

\end{tabular}
}
    \caption{Tracking performance of increasing speed circle trajectories evaluated in real-world scenario~(i).}
    \label{tab:FlightArenaIncreasingSpeedCircleTable}
\end{table}

\begin{figure}[!t]
    \centering
    \vspace{0.7em}
    \includegraphics[width=3.5in,trim={0 20 0 20}]{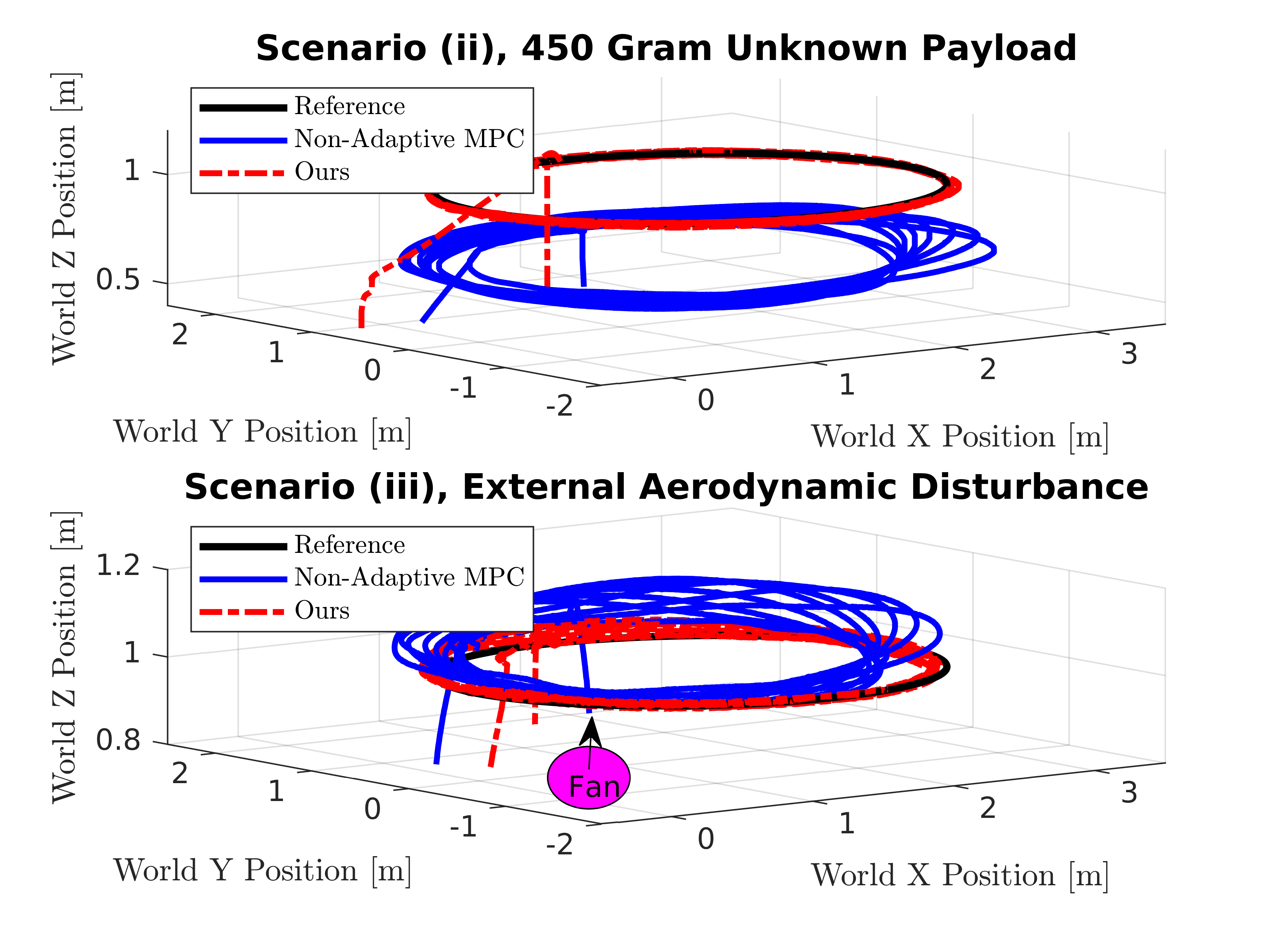}
    \caption{Scenarios (ii) and (iii): Slow speed circle trajectory with induced disturbances. 
    (Top) A payload of \SI{450}{\gram} is added to the quadrotor without updating the nominal NMPC model. 
    The adaptive controller learns the disturbance and immediately compensates for it. 
    (Bottom) An external aerodynamic disturbance is added in the form of a fan and force the quadrotor to fly directly through the turbulence generated from the fan. The adaptive controller compensates immediately for the disturbance and keep the quadrotor close to the reference trajectory.}
    \label{fig:SlowSpeedCircle_FlyingRoom}
\end{figure}

\begin{figure}[t]
    \centering
    \includegraphics[width=0.95\linewidth,trim={20 5 20 10},clip]{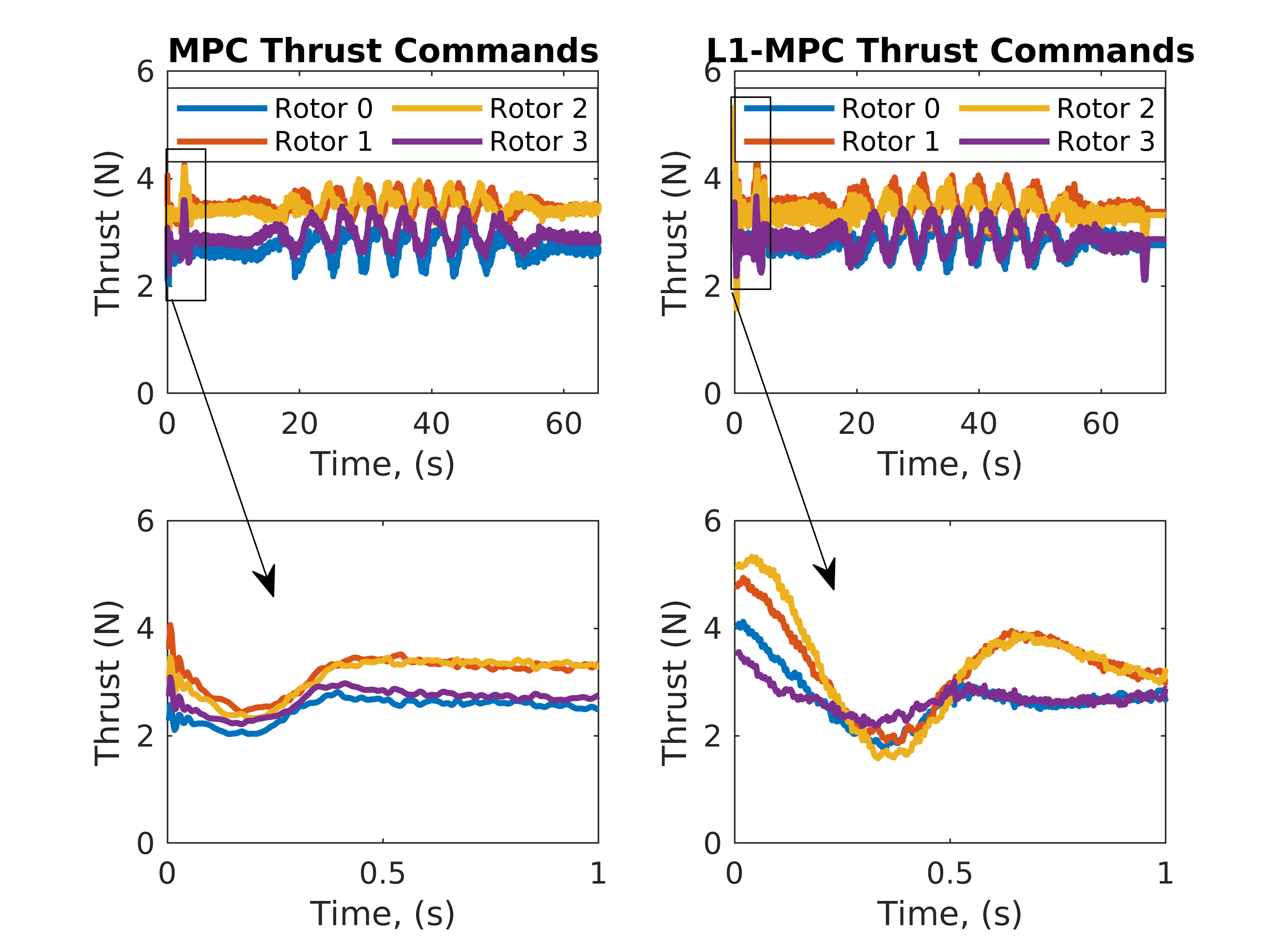}

    \caption{Thrust commands for Scenario (ii). (Top Left) MPC Commands. (Top Right) $\mathcal{L}_1$-MPC Commands. (Bottom Left) First second of thrust commands from MPC, resulting in large steady state offset in Z-position. (Bottom Right) First second of thrust commands from $\mathcal{L}_1$-MPC resulting in immediate adaptation to the unknown payload.}
    \label{fig:Thrust}
\end{figure}

Next, in Setting~(ii) we perform a \SI{2}{\meter\per\second} circle with a \SI{450}{\gram} payload attached.
This represents a mass increase of over 60\%.
The MPC model is not updated to reflect the changed mass and inertia of the system and the adaptive controller must compensate for a large disturbance from takeoff. 
Fig.~\ref{fig:SlowSpeedCircle_FlyingRoom} shows the tracking performance of the quadrotor for scenarios (ii) and (iii).
Relative to the non-adaptive baseline, the $\mathcal{L}_1$-NMPC rejects the disturbance and accurately tracks the specified trajectory.
It is important to note that the adaptation to the unknown payload takes place rapidly, resulting in less than \SI{1}{\cm} steady state Z position tracking error.
In fact, the adaptive controller immediately compensates for the unknown payload by applying higher thrust magnitudes to each rotor upon takeoff, as can be seen in Figure~\ref{fig:Thrust}.
In contrast, the non-adaptive MPC has over a \SI{35}{\cm} steady state Z position error and cannot compensate for the modified inertia of the platform, resulting in overshoots straying away from the reference.

In Setting~(iii) we fly the same trajectory without payloads but instead add an external aerodynamic disturbance in the form of a powerful fan which the quad has to fly directly in front of.
The baseline NMPC immediately demonstrates sizeable tracking performance degradation, while the $\mathcal{L}_1$-NMPC is able to compensate for these disturbances in real time and maintain acceptable, albeit reduced performance. 
In addition to being able to compensate for large parametric disturbances, the $\mathcal{L}_1$-NMPC substantially reduces the tracking error when an exogenous disturbance is introduced.
We believe this has large implications for quadrotors conducting both inspection and shipping tasks outdoors in windy environments.

For Settings (iv) and (v), we fly a highly dynamic race track trajectory generated using the method described in~\cite{foehn2021cpc}.
We compare our proposed approach against the Non-Adaptive MPC and INDI-NMPC in these settings due to the latter's demonstrated success on aggressive racing trajectories.
In Setting (iv), we attach an unknown \SI{100}{\gram} slung payload to introduce a time varying disturbance and fly a mildly aggressive trajectory with peak velocities of \SI{11.9}{\meter\per\second} and peak linear accelerations of over 2 g's.
With the unknown slung load attached, the $\mathcal{L}_1$-NMPC controller reduces tracking error by 44\% relative to the Non-Adaptive NMPC \emph{without} a payload attached. 
Additionally, we outperform INDI-NMPC by 34\% in tracking error, and show that the performance of $\mathcal{L}_1$-NMPC with and without payloads are nearly identical. 
This indicates that the adaptive controller is driving the true system dynamics towards those specified by the reference system, regardless of present disturbances.

\begin{figure*}[h]
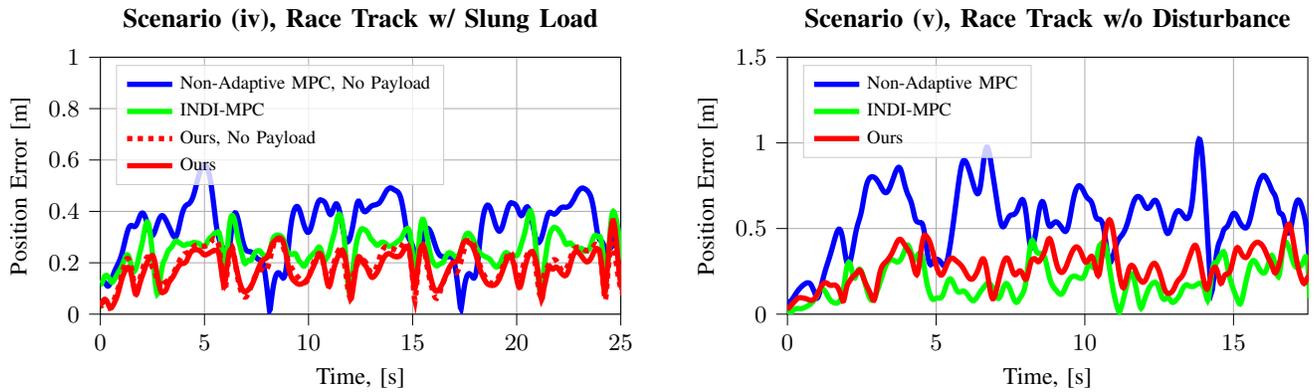

\centering
\input{figures/tikz/fig6_left}%
\hspace{0.5cm}%
\input{figures/tikz/fig6_right}%
\caption{Scenarios (iv) and (v): Tracking error over the course of a dynamic race track. The adaptive controller enforces the quadrotor to fly just as the NMPC model describes it should, regardless if there is a payload present. We demonstrate that when an unknown payload is attached, the $\mathcal{L}_1$-NMPC controller outperforms the non-adaptive baseline \emph{without} a payload attached on an agile racing trajectory with peak speeds of \SI{11.9}{\meter\per\second} (left), and can achieve high tracking accuracy on extremely agile trajectories with peak velocities of \SI{19.4}{\meter\per\second} without the need to update controller gains or model parameters (right).}
\label{fig:Scenario4&5}
\end{figure*}

In Setting (v), we do not introduce disturbances but increase the aggressiveness of the trajectory. %
The drone experiences over 4 g's of linear acceleration, and experiences peak linear velocities of around \SI{20}{\meter\per\second}.
The tracking performance for each of these scenarios can be seen in Figure~\ref{fig:Scenario4&5}.
The $\mathcal{L}_1$-NMPC shows a performance improvement of 49\% of relative to the baseline SRT-NMPC but is outperformed by INDI-NMPC by less than \SI{5}{\centi\meter} RMSE. The INDI-NMPC controller is a highly capable racing controller, but relies on an accurate aerodynamic model to achieve such levels of performance.
With our proposed method we do not require such an accurate model to achieve high tracking accuracy as demonstrated in Scenario (i).
Beyond this, INDI-NMPC requires that the model be updated anytime there is a mass change in order to maximize performance, whereas our controller can learn and compensate the disturbance without the need of updating model parameters or controller gains. 
Across all of these experiments no adjustments are made to the high level MPC, nor the adaptive controller. 
Even with this constraint, we are able to accurately track a variety of trajectories with various unknown disturbances. 
This demonstrates substantial flexibility and robustness when deploying the quadrotor in uncertain scenarios for minimal computational overhead.

\section{Discussion}\label{sec:discussion}
Through extensive experimentation, the $\mathcal{L}_1$ adaptive controller demonstrably enhances a baseline model predictive controller by enabling immediate compensation for model mismatch and external disturbances. 
Cascading the baseline NMPC with an adaptive controller provides substantial benefits in tracking performance across all test cases, especially when there are model mismatches present. 
The performance benefit from the adaptation law is at least 50\% greater than the data driven model augmentation provided by GP-MPC, without the large computational overhead.
When implementing $\mathcal{L}_1$-MPPI, we saw the same high frequency oscillatory content in the body rates that is evident in the videos released with~\cite{pravitra2020l1adaptive}.
While MPPI has shown outstanding results on ground vehicles~\cite{williams2017MPPI,williams2017information}, we believe the sampling approach suffers from the curse of dimensionality problem induced by the 4D space.
We conclude this discussion with the following recommendation: If a user requires aggressive maneuvers and accurate tracking on trajectories with over 4g linear accelerations and there is high confidence in the model accuracy, INDI-NMPC offers the best performance over all controllers considered for onboard computation.
In fact, INDI-NMPC provides the highest autonomous quadrotor racing performance of any control architecture previously studied in the literature to date as indicated by the results in this study and the results in~\cite{sun2021comparative}.
On the other hand, if the user is operating in uncertain environments which may cause large disturbances and require transportation of various payloads without the ability to update the model, $\mathcal{L}_1$-NMPC offers over a 90\% performance improvement over non-adaptive NMPC and INDI-NMPC.
Like the non-adaptive NMPC, INDI-NMPC does not offer any integrative action on linear accelerations and therefore will always have non-zero steady state error in Z-height tracking when payloads are present. 
This is evidenced by the results in Scenario (iv) where we showed our proposed $\mathcal{L}_1$-NMPC outperforming INDI-NMPC when an unknown slung payload is introduced into the system.

\section{Conclusion}\label{sec:conclusion}
In this work, we proposed augmenting a multi-shooting NMPC with an $\mathcal{L}_1$ adaptive inner-loop controller to compensate for model mismatch which can significantly degrade the performance of the NMPC.
We derived the adaptation law at the individual rotor thrust level and leveraged the same nonlinear dynamics model that is embedded in the model predictive controller. The adaptive controller demonstrated minimal computational overhead, and can compute corrective control signals in only 10 microseconds which drastically improves the system performance relative to the non-adaptive NMPC baseline. 
We showed that our proposed method can outperform state of the art data driven MPC methods in both simulation and extensive real world testing on the quadrotor tracking problem.
Instead of having to retrain data driven models online, we instead simply adapt in real time to both parametric and non-parametric disturbances.
Finally, we have shown that the proposed $\mathcal{L}_1$-NMPC can fly highly aggressive racing trajectories while also being able to carry payloads with no additional gain tuning.
One drawback of our proposed approach is the potential for violating actuator constraints due to the inner loop cascade. 
In the future, we plan to explore using the uncertainty estimation from the adaptation law to update our nominal dynamics model online.
We hypothesize that this could significantly improve the prediction accuracy of the NMPC, thus enabling a highly adaptive optimal controller which can fully obey state and input constraints.

\section{Acknowledgment}
The authors thank Leonard Bauersfeld for taking the image used in Figure \ref{fig:Nearmiss} and his advice on editing the accompanying video. Additionally, the authors thank Thomas Laengle for his hardware troubleshooting in the early stages of this project.

{\footnotesize
\bibliographystyle{IEEEtran}
\balance
\bibliography{references}
}

\end{document}